\documentclass{article}
\usepackage{ijcai21}

\usepackage{times}
\usepackage{soul}
\usepackage{url}
\usepackage[hidelinks]{hyperref}
\usepackage[utf8]{inputenc}
\usepackage[small]{caption}
\usepackage{graphicx}
\usepackage{amsmath}
\usepackage{amsthm}
\usepackage{booktabs}
\usepackage[table]{xcolor}
\usepackage{color, colortbl}
\definecolor{LightCyan}{rgb}{0.99, 0.98, 0.65}
\newcolumntype{a}{>{\columncolor{LightCyan}}c}
\usepackage[ruled,vlined]{algorithm2e}

\usepackage{xcolor}

\title{GraphVICRegHSIC: Towards improved self-supervised representation learning for graphs with a hyrbid loss function}

\author{
    Sayan Nag
    \affiliations
    University of Toronto
    \emails
    nagsayan112358@gmail.com
}

\begin{document}

\definecolor{commentcolor}{RGB}{0, 128, 128} 

\definecolor{greencolor}{RGB}{34,139,34} 

\definecolor{bluecolor}{RGB}{0,0,205} 

\newcommand{\PyComment}[1]{\ttfamily\textcolor{commentcolor}{\# #1}}  

\newcommand{\funcdef}[1]{\ttfamily\textcolor{greencolor}{#1}} 

\newcommand{\funcname}[1]{\ttfamily\textcolor{bluecolor}{#1}} 

\newcommand{\PyCode}[1]{\ttfamily\textcolor{black}{#1}} 

\maketitle

\begin{abstract}
  Self-supervised learning and pre-training strategies have developed over the last few years especially for Convolutional Neural Networks (CNNs). Recently application of such methods can also be noticed for Graph Neural Networks (GNNs) . In this paper, we have used a graph based self-supervised learning strategy with different loss functions (Barlow Twins\cite{BTpaper}, HSIC\cite{HSICpaper}, VICReg\cite{VICRegpaper}) which have shown promising results when applied with CNNs previously. We have also proposed a hybrid loss function combining the advantages of VICReg and HSIC and called it as VICRegHSIC. The performance of these aforementioned methods have been compared when applied to 7 different datasets such as MUTAG, PROTEINS, IMDB-Binary, etc. Experiments showed that our hybrid loss function performed better than the remaining ones in 4 out of 7 cases. Moreover, the impact of different batch sizes, projector dimensions and data augmentation strategies have also been explored.
\end{abstract}

\section{Introduction}

A mainstream way of training a deep neural network is to feed some input in order to get certain desirable outputs. Such a training is better known as supervised learning in which a sufficient amount of input data and label pairs are fed to the model depending on its complexity. However, such a method calls for a large number of labels thereby rendering the supervised learning scheme practically inapplicable in many applications because of the unavailability of such large annotated datasets. Dataset labeling not only comes with huge overhead costs, but also can be biased depending on certain circumstances. Thus a need for a better learning strategy led to the advent and thereby adoption of self-supervised learning (SSL) which enables the training of models on unlabeled data. This removes the burden of labeling and therefore it is an efficient and lucrative approach. In the presence of a handful of labeled data, using SSL strategy representations from the remaining unlabeled data can be learnt. Now, the same can be used as a pre-training procedure after which labeled data are used to fine-tune the pre-trained models. These fine-tuned models can either be used for downstream tasks, or as an auxiliary training task that enhances the performance of main tasks. SSL has shown promising results in the field of computer visions using cnns \cite{simsiam,swav,seer,dino,byol,moco,SIMCLRpaper,cnnsurvey} as well as with graph neural networks \cite{zhu2021graph44,hassani2020contrastive9,thakoor2021bootstrapped46,zhu2020deep45,velickovic2019deep37,peng2020graph41,sun2019infograph,SIMCLRpaper,gnnsurvey}.

The primary contributions of this paper are:

\begin{enumerate}

\item We have proposed a hybrid VICRegHSIC Loss function and compared it with five different loss functions namely InfoNCE, SimCLR, Barlow Twins Loss, HSIC Loss and VICReg Loss in the Graph based SSL paradigm and have compared their performances on node classification tasks corresponding to 7 different datasets. Results show that our proposed loss function performs better than the state-of-the-art contemporaries in the framework of Graph SSL. Furthermore, our self-supervised representation learning framework is symmetric as opposed to state-of-the-art frameworks like BGRL (citation).

\item We have conducted ablation studies to demonstrate the impacts of different batch sizes, graph data augmentations and projector dimensions on the proposed loss function.

\item We have also explored the impact of different coefficients (hyper-parameters) of the proposed loss function on the node classification performance tasks.

\end{enumerate}

\begin{figure}
    \centering
    \includegraphics[width = 0.5\textwidth]{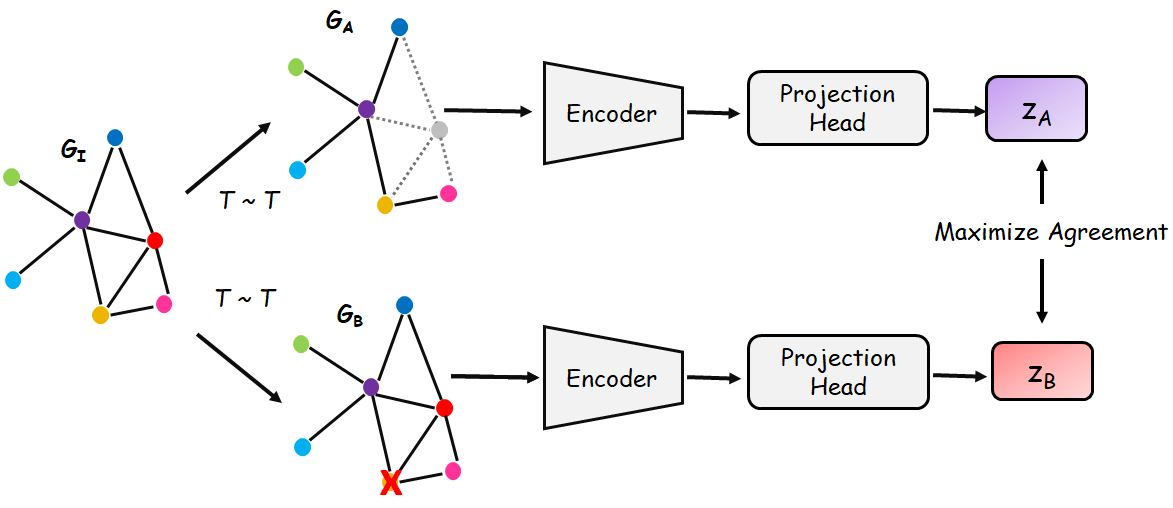}
    \caption{A standard paradigm of graph based self-supervised learning. Graph augmentations on input graph $G_{I}$ resulted in two transformed graphs $G_{A}$ (via node-dropping augmentation) and $G_{B}$ (via attribute-masking augmentation) following each transformation $T$. A self-supervised loss is then computed between the encoded representations $z_A$ and $z_B$ in order to maximize the agreement.}%
    \label{fig:fig1}
\end{figure}

\section{Related Works}

Self-supervised learning has gained a lot of momentum nowadays. It is being applied more in computer vision tasks \cite{cnnsurvey}. Barlow Twins is one such self-supervised learning framework which tries to maximize the agreement between two learned representations of a pair of distorted inputs with a novel Barlow Twins loss function having an invariance term and a redundancy reduction term \cite{BTpaper}. Using Hilbert-Schmidt Independence Criterion, the authors in \cite{HSICpaper} introduced a modification to Barlow Twins Loss by altering the off-diagonal component of the empirical cross-correlation matrix. Using such a negative-sample-free contrastive approach, the authors claimed that the representations learnt will be superior. VICReg \cite{VICRegpaper} uses the Barlow Twins' mechanism of decorrelation without any normalization in its loss function. Furthermore, it can prevent dimension collapse with the variance regularization term.

On the other hand contrastive methods like SimCLR \cite{SIMCLRpaper} use positive and negative sample pairs in their frameworks. More can be found in the survey paper \cite{cnnsurvey}. Self-supervised learning has also been implemented for graph neural netorks. Methods like \cite{velickovic2019deep37,peng2020graph41,sun2019infograph} use contrasting views between nodes and graph whereas \cite{zhu2021graph44,hassani2020contrastive9,thakoor2021bootstrapped46,zhu2020deep45} use contrasting views between nodes and subgraphs and structurally transformed graphs. \cite{qiu2020gcc47} uses contrasting views among subgraphs whilst \cite{graphCLpaper} used such contrasting views between subgraphs and transformed graphs via random augmentations. \cite{gnnsurvey} provides an excellent review of such state-of-the-art methods. Very recently and almost concurrently a Graph SSL framework using Barlow Twins loss function has been proposed in (citation) where the authors used only two augmentations including Node Masking and Edge Dropping. Furthermore, they only focused on a full-batch scenario which they have claimed as one of their limitations. In our work, we have shown the impact of batch sizes and projector dimensions on not only our proposed function but also for Barlow Twins, HSIC and VICReg losses for the MUTAG dataset. Furthermore, we have used four different graph data augmentations namely Edge Dropping, Node Dropping, Node Masking and Subgraphs. Overall, the field is rapidly advancing with promising results and we hope to witness more such advancements in the future.

\section{Methods}

A standard procedure for self-supervised learning for graphs includes graph data augmentation in order to create a pair of graphs, then finding the representations using a graph encoder followed by a projection head and finally using a loss function to maximize agreement between the pair of representations. The aforesaid paradigm has been represented in the Figure \ref{fig:fig1} where the projection head is not shown separately and can be considered to be existing in tandem with the encoder. The method sections comprises of two primary sub-sections, namely, the data augmentation strategies as well as the Self-Supervised Learning framework.

\subsection{Data Augmentation Methods for Graphs}

In data analysis studies, data augmentation is a well-established technique which aims at increasing the amount of data by applying certain transformation strategies on the already existing data without modifying the semantics label. In standard machine learning practices, it is used to reduce over-fitting of the model, thus enhancing the generalizability of the model during predictions. For, graphs, the authors in \cite{graphCLpaper} have explored certain graph-level augmentations which are summarized in the table \ref{tab:data_aug}. For fair comparisons, in our work we have followed the augmentations used by the authors which are explained as follows:

\begin{enumerate}
    \item \textbf{Node Dropping}: For a certain graph, certain portion of vertices can be randomly dropped along with their associated connections. The missing part of vertices, however, does not alter the semantics of the graph.
    
    \item \textbf{Subgraph}: A subgraph is sampled from the main graph using a random walk algorithm \cite{graphCLpaper} with the underlying assumption that the local structure is able to hint the full semantics.
    
    \item \textbf{Edge Perturbation}: The connectivities of edges are perturbed by randomly introducing or discarding a certain proportion of them. The underlying prior in this case is that the graph has a certain degree of semantic robustness to the connectivity variations.
    
    \item \textbf{Attribute Masking}: In this case, vertex attributes are masked to ensure that the model recovers those masked vertex attributes using the neighbouring structure assuming that there exists a certain extent of semantic robustness against losing partial attributes.
    
\end{enumerate}

The influences of the node-dropping, subgraph, edge-perturbation and attribute-masking factors on the performance of the algorithm will be shown in the results section.

\begin{figure*}[!h]
    \centering
    \includegraphics[width=\textwidth, height = 0.25\textwidth]{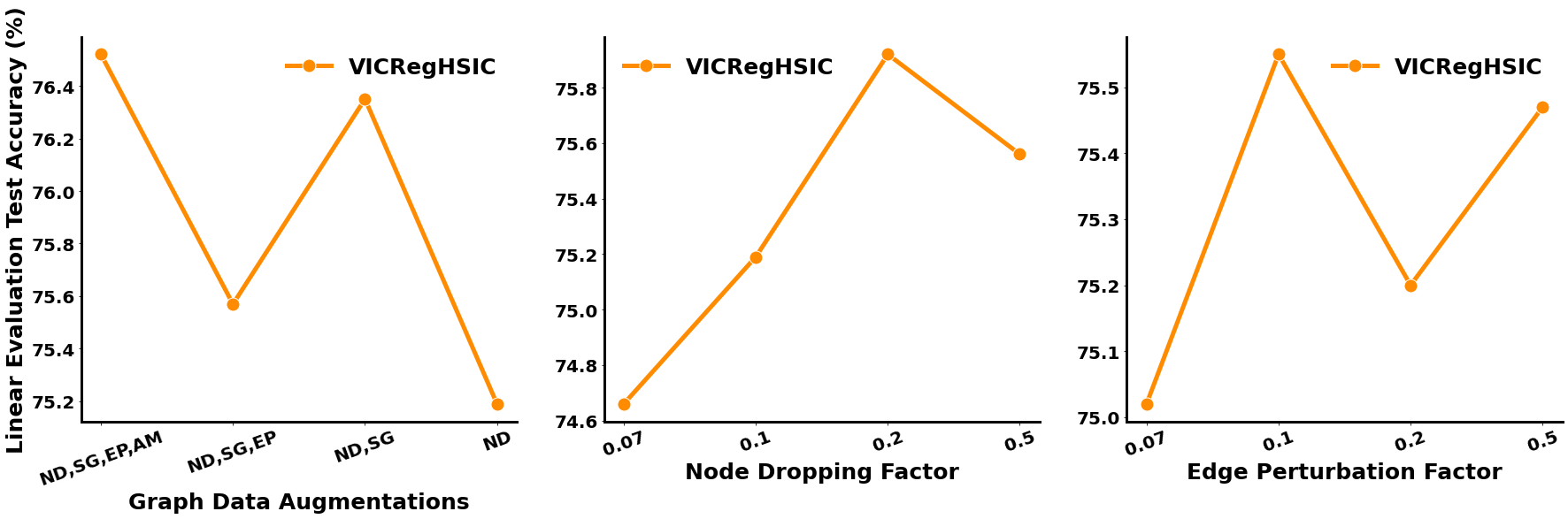}
    \includegraphics[width=\textwidth, height = 0.25\textwidth]{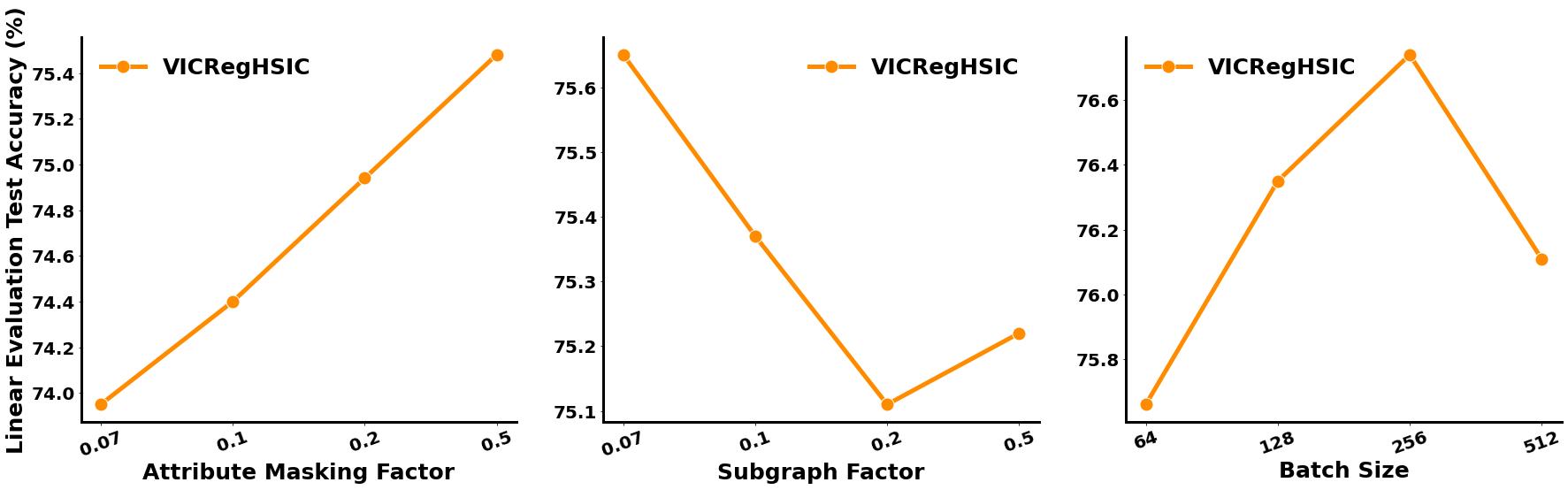}
    \caption{Impact of different graph data augmentations and batch-sizes on linear evaluation test accuracy on PROTEINS}%
    \label{fig:Pro_aug_bs}%
\end{figure*}

\subsection{Framework}

In this section a graph self-supervised learning framework is proposed for self-supervised pre-training of GNNs. In such a paradigm, the agreement between an augmented pair of representations of the same graph is maximized using a loss function as illustrated in the Figure \ref{fig:fig1}. The entire paradigm is composed of four parts as follows:

\begin{enumerate}
    \item \textbf{Data Augmentation}: For a certain graph, two augmented versions are obtained as a pair after applying transformations on the actual graph.
    
    \item \textbf{Encoder}: A graph encoder is used to extract a pair of graph-level representations for the corresponding pair of augmented graphs.
    
    \item \textbf{Projection Head}: A non-linear projection head (a two-layer perceptron in this case) is used to map the pair of augmented graph-level representations to a latent space for applying the loss function.
    
    \item \textbf{Loss Function}: A self-supervised loss function is used to maximize the agreement between the pair of latent representations obtained as outputs from the projection head. In our work we have proposed a negative-sample free hybrid loss function named VICRegHSIC loss, combining the VICReg loss \cite{VICRegpaper} and HSIC loss \cite{HSICpaper}. In the hybrid loss we have just modifed the on- and off-diagonal terms of the covariance component of the VICReg as explained in \cite{HSICpaper}. Apart from the aforementioned proposed loss function, for comparison purposes we have used other different self-supervised and contrastive loss functions such as Barlow-Twin, Hilbert-Schmidt Independence Criterion, Variance-Invariance-Covariance Regularization, SimCLR, InfoNCE losses among others.
    
    \subsubsection{Variance-Invariance-Covariance Regularization}
    
    Variance-Invariance-Covariance Regularization Loss (VICReg) has been proposed recently as an improvement over the existing Barlow Twins loss function. The authors in \cite{VICRegpaper} have claimed that VICReg is a simple yet effective loss which prevents information collapse in self-supervised joint embedding learning paradigms leading to better training stability and improvement in performance on several downstream tasks. In this paper, we have extended that concept to graph-structured data where we have used a modified version of the loss for downstream task such as graph node classification. Considering \textit{$Z_A$} and \textit{$Z_B$} as the embeddings (outputs from the projection head, refer to the Figure \ref{fig:fig1}) we can compute the following losses:
    
    \textbf{Variance Regularization} loss is represented in the form of a hinge function on the regularized standard deviation \textit{S} of the embeddings along a batch dimension \textit{d} as:
    \begin{gather}
        v(Z_A) = \frac{1}{d}\sum_{j=1}^{d}max(0,\gamma - S(z_A^j,\epsilon)),\\
        S(x,\epsilon) = \sqrt{Var(x) + \epsilon}
    \end{gather}
    As described in \cite{VICRegpaper}, $\epsilon$ is a small scalar which prevents numerical instabilities and $\gamma$ is the constant target value for the standard deviation which is fixed to 1 for all the experiements that we have conducted.
    
    \textbf{Covariance Regularization} loss (\textit{c}) is given as the sum of the squared off-diagonal coefficents of the Covariance matrix (\textit{C}) scaled by the dimension (\textit{d}):

    \begin{gather}
        C(Z_A) = \frac{1}{n-1}\sum_{i=1}^{n}(z_{Ai} - \tilde{z}_A)(z_{Ai} - \tilde{z}_A)^T,\\
        \tilde{z}_A = \frac{1}{n}\sum_{i=1}^{n}z_{Ai},\\
        c(Z_A) = \frac{1}{d}\sum_{i \ne j}[C(Z_A)]_{i,j}^{2}
    \end{gather}
    
    \textbf{Invariance} is defined as the mean squared p-norm distance between each pair of vectors without any normalization given as:
    
    \begin{gather}
        s(Z_A,Z_B) = \frac{1}{n}\sum_{i}\|z_{Ai} - z_{Bi}\|_{p}^{2} 
    \end{gather}
    
    In the actual paper \cite{VICRegpaper}, the authors have considered $p = 2$, but have considered a general case and have done ablations with 4 different p values and have found $p = 2$ works the best for us (see Section 4, Experiments and Ablations).
    
    The final loss function is a weighted sum of the individual variance, invariance and covariance regularization terms given as:
    
    \begin{equation}
        \begin{split}
            l(Z_A,Z_B) = \lambda s(Z_A,Z_B) + \mu [v(Z_A) + \\ v(Z_B)] + \nu [c(Z_A) + c(Z_B)]
        \end{split}
    \end{equation}
    
    Here, $\lambda$, $\mu$, and $\nu$ are the hyper-parameters which weight the individual regularization terms. As outlined in the \cite{VICRegpaper}, we have considered $\nu = 1$ . For $\lambda$ and $\mu$ we have done a grid-search and some ablation studies as outlined in the later section of the paper. $\lambda = \mu = 25$ seemed to work the best in our case.
    
    \subsubsection{Hilbert-Schmidt Independence Criterion}
    
    The representation vectors in the Barlow Twins method are presumed to follow a Gaussian distribution. However, this presumption can be relaxed and a bridge between between the contrastive and non-contrastive learning can be constructed using a negative sample-free contrastive learning approach using the so-called Hilbert-Schmidt Independence Criterion (HSIC) \cite{HSICpaper}. The HSIC approach can be employed in the covariance regularization term and can be suitably modified as:
    
    \begin{gather}
        c(Z_A) = \frac{1}{d}\sum_{i \ne j}[1 + C(Z_A)_{i,j}]^{2}
    \end{gather}
    
    For a detailed implementation of the proposed VICRegHSIC loss, please refer to Algorithm \ref{algo:VICRegHSIC}.
    
\end{enumerate}

\begin{algorithm*}[!ht]\footnotesize
\SetAlgoLined
    \PyComment{f: Graph Encoder} \\
    \PyComment{$\lambda$, $\mu$, $\nu$: coefficients of invariance, variance and covariance losses} \\
    \PyComment{N: batch size} \\
    \PyComment{D: representation dimension} \\
    \PyComment{mseloss: Mean square error loss} \\
    \PyComment{offdiagonal: off-diagonal elements of a matrix} \\
    \PyComment{relu: ReLU activation function} \\
    \PyComment{G: Actual Graph} \\
    \PyComment{augment: augmentation function} \\
    \PyComment{VICRegHSIC: proposed self-supervised loss function} \\
    \PyCode{\funcdef{def} \funcname{VICRegHSIC}(G, augment)} \\
    \Indp   
        \PyComment{load a batch with N samples} \\
        \PyCode{\funcdef{for} x in loader:} \\
        \Indp
            \PyCode{$G_A$, $G_B$ = augment(G)} \PyComment{ two randomly augmented versions of G}\\
            \PyCode{$z_A$ = f($G_A$)} \PyComment{N x D representation}\\
            \PyCode{$z_B$ = f($G_B$)} \PyComment{N x D representation}\\
            \PyCode{simloss = mseloss($z_A$, $z_B$)}\PyComment{invariance loss}\\
            \PyCode{stdzA = torch.sqrt($z_A$.var(dim=0) + 1e-04)}\\
            \PyComment{variance loss}\\
            \PyCode{stdzB = torch.sqrt($z_B$.var(dim=0) + 1e-04)}\\
            \PyCode{stdloss = torch.mean(relu(1-stdzA)) + torch.mean(relu(1-stdzB))}\\
            \PyComment{covariance loss}\\
            \PyCode{$z_A$ = $z_A$ - $z_A$.mean(dim=0)}\\
            \PyCode{$z_B$ = $z_B$ - $z_B$.mean(dim=0)}\\
            \PyCode{covzA = ($z_A$.T @ $z_A$)/(N - 1)}\\
            \PyCode{covzB = ($z_B$.T @ $z_B$)/(N - 1)}\\
            \PyCode{covloss = offdiagonal(covzA).add\_(1).pow\_(2).sum()/D}\\ \PyCode{covloss += offdiagonal(covzB).add\_(1).pow\_(2).sum()/D}\\
            \PyComment{loss}\\
            \PyCode{VICRegHSICloss = $\lambda$ * simloss + $\mu$ * stdloss + $\nu$ * covloss}\\
        \Indm
        \PyCode{return VICRegHSICloss}\\
    \Indm 
\caption{PyTorch-style pseudocode for VICRegHSIC}
\label{algo:VICRegHSIC}
\end{algorithm*}

\begin{table}[!ht]\footnotesize
    \centering
    \renewcommand{\arraystretch}{1.3}
    \begin{tabular}{c|c}

    \hline
    Augmentation Methods & Type\\[.5ex]
    \hline
    \hline
    Node Dropping (ND) & Nodes\\[.5ex]
    
    Subgraph (SG) & Nodes, Edges\\[.5ex]
     
    Edge Perturbation (EP) & Edges\\[.5ex]
    
    Attribute Masking (AM) & Edges\\[.5ex]
    \hline
    \end{tabular}\\[0.5ex]
    \caption{Data Augmentation methods for graphs.}
    \label{tab:data_aug}
\end{table}

\section{Experiments and Ablations}

For our work we have considered 7 different datasets namely MUTAG, PROTEINS, IMDB-Binary, Cox2-MD, PTC-FR, Synthetic and AIDS datasets for node-classification tasks. In this section we will showcase the impacts of various augmentations, loss function hyperparameters, batch sizes as well as dimensions of the projection head. 

\begin{figure*}[!h]
    \centering
    \includegraphics[width=\textwidth, height = 0.25\textwidth]{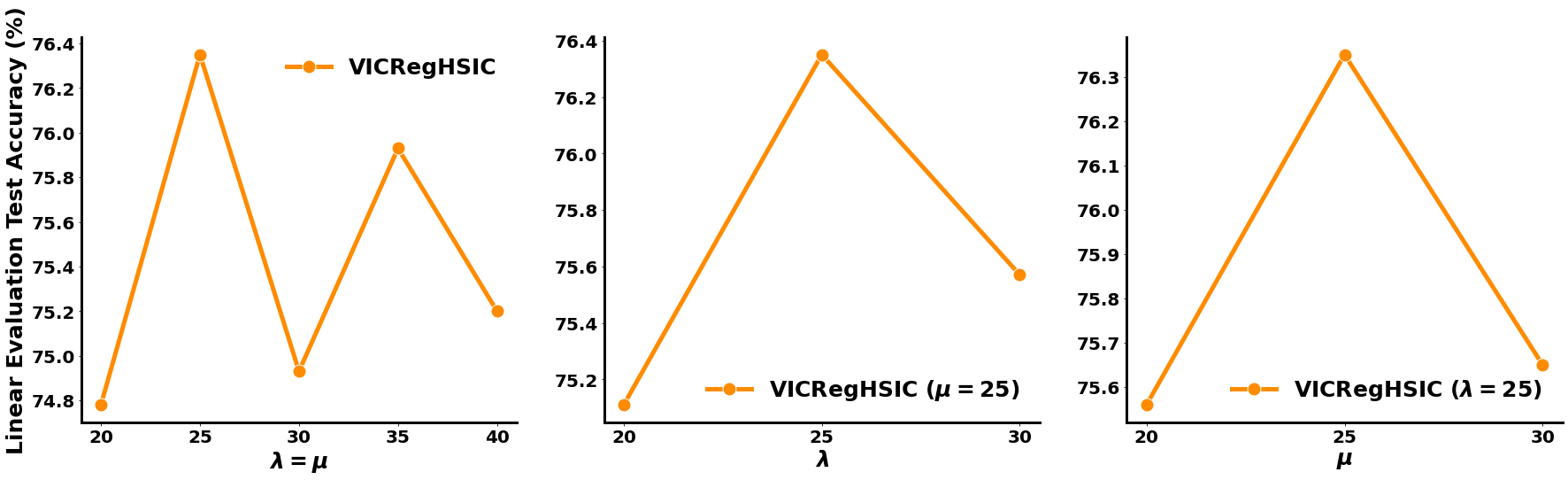}
    \caption{Impact of different $\lambda$ and $\mu$ values (coefficients of the proposed VICRegHSIC loss function) on linear evaluation test accuracy on PROTEINS}%
    \label{fig:Pro_lambda}%
\end{figure*}

\begin{figure*}[h]
    \centering
    {{\includegraphics[width=0.4\textwidth, height = 0.25\textwidth]{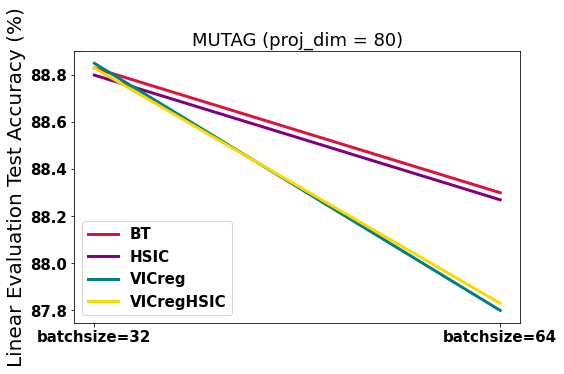} }}%
    \hspace{8mm}
    {{\includegraphics[width=0.4\textwidth, height = 0.25\textwidth]{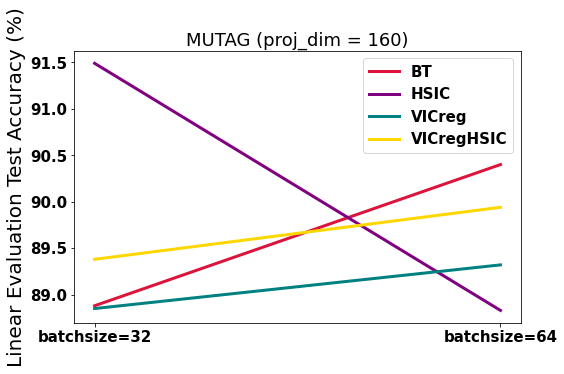} }}%
    {{\includegraphics[width=0.4\textwidth, height = 0.25\textwidth]{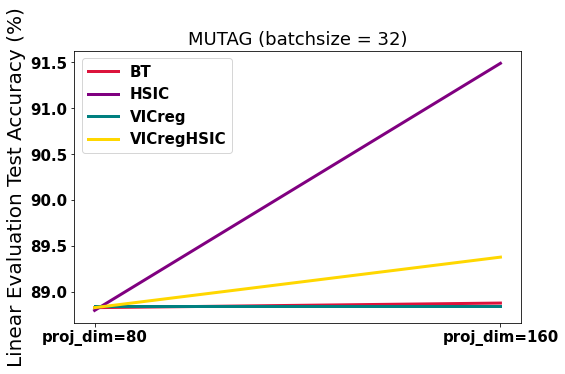} }}%
    \hspace{8mm}
    {{\includegraphics[width=0.4\textwidth, height = 0.25\textwidth]{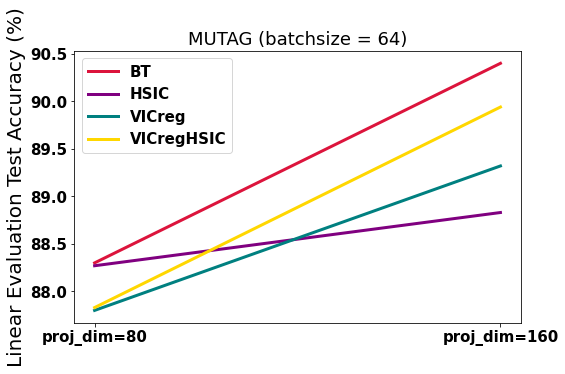} }}%
    \caption{Impact of batch sizes and projector dimensions on linear evaluation test accuracy on MUTAG}%
    \label{fig:PD_BS}%
\end{figure*}

\subsubsection{Impact of Graph Data Augmentations}

The Figure \ref{fig:Pro_aug_bs} shows that for PROTEINS dataset, removing the data augmentations reduces the performance, i.e., with all the four augmentations the performance is the maximum. This means that once all the aforesaid augmentations are applied, it enforces the model to learn representations invariant to the desired perturbations in a better way through minimizing the self-supervised loss among the augmented pairs.

It may be worth noting that edge perturbation leads to a decrease in performance overall for both PROTEINS and MUTAG. This is because for biomolecules datasets, the semantics are less robust against connectivity variations \cite{graphCLpaper}. Edge perturbations in such biomolecule datasets create modifications (addition or removal) of intermediate covalent bonds leading to radical alterations in the identities, composition and properties of chemical compounds. This affects the down-stream performance which is reflected in the above figures.

Node dropping augmentation have been seen to be useful in the biomolecule dataset like PROTEINS (Figure \ref{fig:Pro_aug_bs}). This indicates that missing vertices, for example, certain atoms in the chemical compounds (graphs) does not alter the semantics.

Unlike what has been claimed in \cite{graphCLpaper} subgraph augmentation has reduced the performance of the algorithm drasctially with the increasing in the factor (Figure \ref{fig:Pro_aug_bs}). This indicates that for biomolecule datasets local structure is always not able to hint the full semantics of the chemical compound. On the other hand, increasing the masking factor for attribute masking augmentation leads to an increase in performance confirming the fact that the semantic robustness against losing partial attributes due to masking has been preserved (Figure \ref{fig:Pro_aug_bs}).

\subsubsection{Impact of VICRegHSIC Loss Function Hyperparameters}

In \cite{VICRegpaper}, the authors have considered $\lambda = \mu$ for the coefficients of the loss function. Similarly, in our case, we have considered the same for the rest of the experiments and ablation studies. However, we ran some experiments by varying both $\lambda$ and $\mu$ keeping $\lambda = \mu$ and also $\lambda$ and $\mu$ individually keeping the other constant. Figure \ref{fig:Pro_lambda} shows that $\lambda=\mu=25$ works best for the algorithm for the PROTEINS dataset. For simplicity we have kept $\lambda=\mu=25$ for other experiments and datasets too.

\subsubsection{Impact of Batch Sizes and Projector Dimensions}

Increase in the batchsize as shown in figure \ref{fig:Pro_aug_bs} leads to an increase in accuracy for PROTEINS with a few exceptions. The Figure \ref{fig:PD_BS} shows that for MUTAG dataset increase in batch size for a projector dimension of 80 leads to a decrease in performance. However, increasing the projector dimension from 80 to 160 reverses the effect of increasing batch size and all except HSIC gives better performance. This is consistent with the performance of HSIC as demonstrated in \cite{HSICpaper}. For this analysis, we have considered data augmentations Node Drop and Subgraph together in the augmentation pool.

\begin{table*}[ht]\footnotesize
    \centering
    \renewcommand{\arraystretch}{1.3}
    \begin{tabular}{cccccccc}

    \toprule
    \textbf{Methods} & \textbf{MUTAG} & \textbf{PROTEINS} & \textbf{IMDB-B} & \textbf{COX2-MD} & \textbf{PTC-FR} & \textbf{SYNTHETIC} & \textbf{AIDS}\\[.5ex]
    \midrule
    
    InfoGraph \shortcite{sun2019infograph} & $89.13\pm0.97$ & $74.48\pm0.92$ & $73.05\pm1.33$ & $51.26\pm0.71$ & $64.18\pm0.59$ & $54.86\pm1.25$ & $99.42\pm0.81$\\[.5ex]
    
    GCL \shortcite{graphCLpaper} & $86.98\pm1.03$ & $74.43\pm0.22$ & $72.21\pm0.95$ & $50.15\pm1.93$ & $65.49\pm0.47$ & $55.28\pm2.11$ & $99.62\pm0.44$\\[.5ex]
    
    GBT & $89.41\pm0.93$ & $76.12\pm0.56$ & $72.50\pm0.22$ & $\textbf{54.11}\pm\textbf{0.71}$ & $65.54\pm0.38$ & $57.33\pm1.34$ & $99.68\pm0.36$\\[.5ex]
     
    GHSIC & $88.68\pm0.85$ & $76.08\pm1.06$ & $71.93\pm0.33$ & $53.83\pm0.62$ & $65.80\pm0.35$ & $56.42\pm0.67$ & $99.65\pm0.43$\\[.5ex]
    
    GVICReg & $88.83\pm0.42$ & $\textbf{76.37}\pm\textbf{0.68}$ & $72.9\pm0.65$ & $54.05\pm1.26$ & $\textbf{65.84}\pm\textbf{0.96}$ & $58.31\pm1.66$ & $99.70\pm1.13$\\[.5ex]
    
    \rowcolor{LightCyan}
    
    GVICRegHSIC & $\textbf{90.05}\pm\textbf{0.54}$ & $76.35\pm0.43$ & $\textbf{73.2}\pm\textbf{0.18}$ & $54.09\pm0.94$ & $65.81\pm0.47$ & $\textbf{58.33}\pm\textbf{0.88}$ & $\textbf{99.73}\pm\textbf{0.76}$\\[.5ex]
    \bottomrule
    \end{tabular}\\[0.5ex]
    \caption{Linear Evaluation Test Accuracy scores between state-of-the-art methods.}
    \label{tab:sota_results}
    
\end{table*}

The Figure \ref{fig:PD_BS} shows that for MUTAG dataset increase in projector dimension for both the batch sizes leads to an increase in performance. It is more pronounced in batch size 64 and can be assumed that with even further increase in batch sizes the impact will be more. For this illustration, we have considered data augmentations Node Drop and Subgraph together in the augmentation pool.

\begin{figure*}[!h]
    \centering
    \includegraphics[width=0.4\textwidth, height=0.25\textwidth]{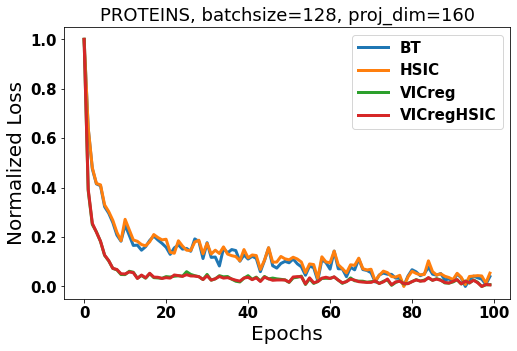}
    \hspace{8mm}
    \includegraphics[width=0.4\textwidth,height=0.25\textwidth]{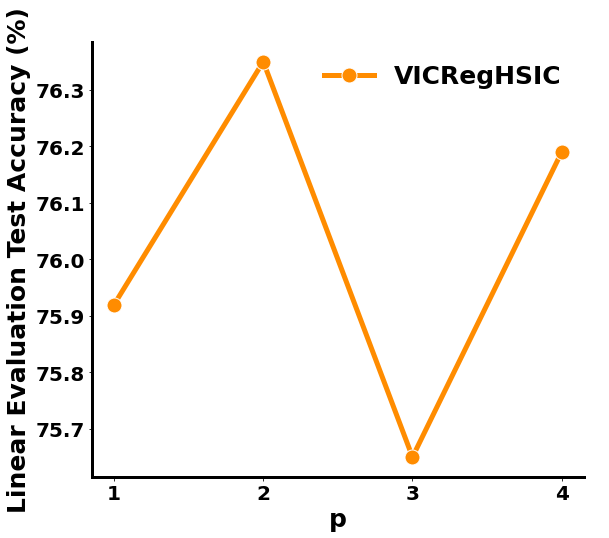}
    \caption{Left: Normalized training loss curves (for representation purposes) for 100 epochs showing performance of different methods on PROTEINS dataset. Right: Performance of invariance term (Lp norm) for different values of p.}%
    \label{fig:lossfn}%
\end{figure*}

\subsubsection{VICRegHSIC Loss: Lp-Norm and Convergence}

The Figure \ref{fig:lossfn} demonstrates the performance of different methods on the PROTEINS dataset when trained for 100 epochs. Faster convergence is observed for VICReg and VICRegHSIC compared to BT and HSIC. The training loss curves have been normalized for representation and comparison purposes. Furthermore, we have explored the possibilities of different p values for the Lp-norm in the invariance term of the proposed loss function (see Figure \ref{fig:lossfn}) and have found $p=2$ to work the best in most of the datasets.

\subsubsection{Performance on Different Datasets}

Linear Evaluation Test Accuracy scores between different state-of-the-art methods with respect to Graph SSL for 7 different datasets have been shown in the Table \ref{tab:sota_results}. For the experiments outlined in the aforementioned table, we have considered a batch size of 128 and projector dimension of 160. For both VICReg loss and the VICRegHSIC loss, we have considered $\lambda=\mu=25$. The above experiments were run 5 times on NVIDIA T4 with 16GB RAM and we have reported the mean and standard deviations. The Table \ref{tab:sota_results} shows that the proposed VICRegHSIC loss function gives the best accuracy scores in 4 out of 7 cases performing marginally better than VICReg Loss.

\section{Conclusion and Future Works}

In this paper, we have studied the performances of different loss functions for self-supervised pre-training of Graph Neural Networks and have compared them with our proposed hybrid VICRegHSIC loss function. Our VICRegHSIC framework is a negative sample-free contrastive learning approach which combines a state-of-the-art VICReg Loss (previously used for Self-Supervised Learning on CNNs for vision related downstream tasks) and the Hilbert-Schmidt Independence Criterion (HSIC). This acts as a bridge between the contrastive and non-contrastive self-supervised learning methods.

In our proposed graph supervised learning framework for GNN pre-training we have compared state-of-the-art self-supervised contrastive and non-contrastive loss functions to facilitate robust representation learning invariant to perturbations via agreement maximization between the pair of augmented representations. We have also analyzed the impact of several data augmentation strategies such as node dropping and attribute masking, etc. in our proposed framework and have noticed that considering all the augmentations together gives the best results in terms of generalizability and robustness. We have done ablation studies on batch sizes and projector dimensions showing that increasing projection dimension increases the performance of the algorithms which is consistent with the findings in \cite{BTpaper,VICRegpaper}. We have explored the possibilities of different p values for the Lp-norm in the invariance term of the proposed loss function and have found $p=2$ to work the best in most of the datasets. The values of $\lambda$ and $\mu$ are also explored where we did not necessarily stick to the norm of considering $\lambda=\mu$, however, such equality between the two hyperparameters seemed to work best for the proposed algorithm in most cases. In the future, more experiments need to be conducted to better understand the performance of these algorithms on a variety of other datasets especially including datasets with more nodes and edges. We did not conduct those experiments because of memory constraints. Other downstream tasks such as edge classification can be considered as a potential future work in the realm of graph self-supervised learning which we could not perform because of time constraints. Further modifications of the loss functions, data augmentation strategies as well as the self-supervised framework could lead to more efficient solutions and even better performances in the domain of graph self-supervised learning.


\newpage

\bibliographystyle{named}
\bibliography{ijcai21}

\end{document}